\documentclass[11pt]{article}
\usepackage[preprint]{acl}

\usepackage{times}
\usepackage{latexsym}
\usepackage[T1]{fontenc}
\usepackage[utf8]{inputenc}
\usepackage{microtype}
\usepackage{booktabs}
\usepackage{amsmath}
\usepackage{graphicx}
\usepackage{xcolor}
\usepackage{url}
\usepackage{pgfplots}
\pgfplotsset{compat=1.17}
\usepgfplotslibrary{groupplots}
\usetikzlibrary{positioning}

\title{Text Speeding Itself Up:\\Recency and Restraint in Session-Local Drafting}

\author{Antal van den Bosch \\
  Utrecht University \\
  \texttt{a.p.j.vandenbosch@uu.nl}}

\begin{document}
\maketitle

\begin{abstract}
Speculative decoding lets a cheap draft model propose tokens that a large
language model verifies in a single forward pass. When the verifier keeps a
KV cache, verification is cheap and drafting must be nearly free.
Session-local drafters, which predict from the prompt and the text generated
so far, meet that bar without training. We compare a recency-weighted,
backing-off, confidence-gated session memory against prompt lookup decoding
in its reference form, which copies from the earliest earlier match, and in
a variant that copies from the most recent one, with paired per-prompt
bootstrap intervals throughout. On general, medical, and legal text with a
Mistral-7B verifier, recency accounts for the speed: most-recent lookup
gains up to 0.09 in speedup over earliest-match lookup on 500-token
generations, and session memory's weighting and backoff add nothing
measurable on top, at 30 to 40 times the drafting cost. Energy tells a
different story. On medical and legal text, unselective drafters are 14\%
faster than plain decoding yet use as much energy per token, while drafters
that abstain on weak evidence save 3 to 4 joules per 100 tokens; a
three-token minimum match length on prompt lookup matches the
confidence-gated session memory. Outputs are token-identical to greedy
decoding in 94\% of generations, with every divergence at a numerical
near-tie. Under a repetition penalty, long-generation speedups shrink from
1.4--1.7$\times$ to 1.1--1.2$\times$, but recency still helps and the
length floor is the only drafter that saves energy in both domains.
\end{abstract}

\section{Introduction}
\label{sec:intro}

Speculative decoding~\citep{leviathan2023speculative,chen2023speculative}
lets a cheap draft model propose several tokens that a large verifier checks
in one forward pass. When the verifier keeps a KV cache, so that each new
token attends to stored keys and values instead of re-encoding the sequence,
verification is cheap and the cost of drafting becomes the binding
constraint: a draft token that takes longer to produce than the verification
it saves is a net loss~\citep{vandenbosch2026green}. The field has moved
accordingly, from separate neural draft models toward attached prediction
heads~\citep{cai2024medusa,li2024eagle}, retrieval from a
corpus~\citep{he2023rest}, and lookup in the prompt
itself~\citep{saxena2023prompt}.

Trained draft models that are cheap enough for this regime exist, such as
the CPU-resident memory-based language models
of~\citet{vandenbosch2025olifant}, but they are fixed at training time.
\citet{vandenbosch2026green} report that a memory-based draft model
trained on legal text has 38\% of its drafts accepted on legal input and
5.9\% on medical input, and falls below break-even under a cached
verifier when the domain does not match.

The alternative is a drafter that needs no training data because it learns
from the session itself, from the
prompt and from the tokens the verifier has already accepted. Prompt lookup
decoding~\citep{saxena2023prompt} is the simplest such drafter: it copies the
continuation of an exact n-gram match found earlier in the sequence. We
introduced a richer variant, a \emph{session memory}: a recency-weighted
variable-order n-gram store with longest-match backoff and an entropy-based
confidence gate. On 500-token greedy generations it reaches
1.4--1.7$\times$, though, as we show, mostly by copying the verifier's own
repetition. The question this paper answers
is which of its ingredients do the work, and it does so by comparing
against prompt lookup decoding in two forms: the reference behaviour, which
copies from the \emph{earliest} match, and a one-line change that copies
from the \emph{most recent} match. That comparison isolates recency; adding
session memory on top isolates weighting and backoff; switching the gate on
isolates abstention. All comparisons are paired over prompts against a
same-session baseline with bootstrap intervals, because differences of a
few percent are the object of study.

The answers are not the ones we expected when we built the mechanism, and
they change what is worth recommending:
\begin{itemize}
\item \textbf{Recency is what matters for speed.} Copying from the most
  recent match instead of the earliest one adds $+0.016$ to $+0.027$ to the
  speedup on 100-token general text and $+0.08$ on 500-token generations.
  Recency weighting and backoff add nothing measurable on top: session
  memory and most-recent lookup are within $0.02$ of each other in every
  cell, and lookup drafts at a thirtieth to a fortieth of the cost
  (Section~\ref{sec:speed}).
\item \textbf{Selectivity is what matters for energy.} At draft length 5,
  unselective drafters run about 14\% faster than plain decoding on
  medical and legal text but use as much energy or more, because they raise mean
  power. The confidence gate loses a little speed and saves 2.7--3.9 joules
  per 100 tokens, and a plain order floor on prompt lookup, drafting only on
  matches of three or more tokens, saves the same amount
  (Section~\ref{sec:energy}). The gate's value is abstention on short
  contexts, not confidence, and the floor is the only drafter that still
  saves energy when a repetition penalty suppresses the verifier's
  self-repetition.
\item \textbf{Speculative output is greedy output, up to numerical ties.}
  Across 2,700 generations, 94\% are token-identical to greedy decoding.
  Every divergence sits at a position where the verifier's top two
  candidates differ by at most 0.023 logits, against a median gap of 3.0,
  and it is independent of the drafter and the draft length
  (Section~\ref{sec:agree}). This replaces the ROUGE-based argument of an
  earlier version of this work.
\item \textbf{Long-generation gains are mostly repetition.} Drafting
  value rises with position in a greedy generation, and the gap between
  most-recent and earliest-match lookup widens with it. Under a repetition
  penalty that removes most of the verifier's self-repetition, every
  drafter falls from 1.4--1.7$\times$ to 1.08--1.16$\times$, while recency
  still helps (Section~\ref{sec:curve}).
\end{itemize}

\section{Session-Local Drafters}
\label{sec:system}

\subsection{Speculative decoding with a cached verifier}

A draft model proposes $k$ tokens; the verifier checks them in a single
pass and accepts the longest prefix matching its own greedy choices. Our
verifier decodes greedily over a persistent KV cache with a rejection
fast-path: the first draft token is verified against the logits left over
from the previous step, which are available for free, so a step whose first
token is rejected costs exactly one plain decoding step. The full $k$-token
verification pass is only performed when the first token is accepted, and
on rejection the cache is truncated to the accepted prefix. A draft call
that proposes nothing also costs one plain step; we call this
\emph{abstaining}. Acceptance is decided at token-ID level against the
verifier's greedy choice, so the emitted text is the verifier's greedy
output up to the numerical effect of batched versus single-token
verification, which Section~\ref{sec:agree} measures.

\subsection{Prompt lookup}
\label{sec:pldef}

Prompt lookup decoding~\citep{saxena2023prompt} drafts by copying. Let
$c_{1:T}$ be the prompt followed by every verifier-accepted token. For
$n$ from 3 down to 1, the last $n$ tokens form a query; an earlier position
where the query occurs is a match, and the $k$ tokens that followed it are
the draft. The reference implementation accepts a match only if a full
$k$-token continuation follows it and it does not overlap the query, and
returns the first valid match in index order, that is, the \emph{earliest}
occurrence. If no order matches, it abstains. We reimplemented it and
checked it token for token against the reference function on random
sequences, and use it in four forms (Table~\ref{tab:drafters}):
\emph{earliest} (the reference behaviour), \emph{most recent} (identical,
but returning the last valid match), an \emph{$m$-token floor} (most
recent, with no backoff below order $m$, for $m\in\{2,3\}$), and \emph{agreement} (most
recent; at the matched order, draft only if every valid occurrence
continues with the same token).

\begin{table}[t]
\centering
\small
\setlength{\tabcolsep}{3pt}
\resizebox{\columnwidth}{!}{%
\begin{tabular}{lll}
\toprule
Drafter & Draws on & Abstains if \\
\midrule
Lookup, earliest & first match, $n\le3$ & no full match \\
Lookup, most recent & last match, $n\le3$ & no full match \\
Lookup, $m$-token floor & last match, $m\le n\le3$ & no match $n\ge m$ \\
Lookup, agreement & all matches, top $n$ & successors differ \\
Session memory & decayed counts, $n\le4$ & nothing matches \\
\quad gated ($\beta$=0.5) & same & $\kappa < \beta$ \\
\bottomrule
\end{tabular}}
\caption{The session-local drafters compared. All draw only on the prompt
and verifier-accepted tokens.}
\label{tab:drafters}
\end{table}

\subsection{Session memory}
\label{sec:session}

The session memory is a recency-weighted, variable-order n-gram store built
from the generation context, over the verifier's own SentencePiece
vocabulary. Let $c_{1:T}=c_1,\dots,c_T$ be the tokens produced so far (the
prompt followed by every verifier-accepted token). We use maximum order
$N=4$ and decay $\alpha=0.95$.

\paragraph{Decayed counts.}
For each position $p$ and each order $n\in\{1,\dots,\min(N,p-1)\}$, the
$n$-gram $g=c_{p-n:p-1}$ is a context in which the token $c_p$ was observed.
Each such observation contributes a weight that decays geometrically with its
distance from the current end of the sequence,
\begin{equation}
  C_n(g,t)\;=\!\!\sum_{\substack{2\le p\le T\\ c_{p-n:p-1}=g,\; c_p=t}}\!\!
  \alpha^{\,T-p},
  \label{eq:counts}
\end{equation}
so the most recent occurrence of a context--token pair carries weight
$\alpha^{0}=1$ and each earlier occurrence is discounted by a further factor
$\alpha$. %

\paragraph{Backoff prediction.}
Prediction is based on finding the longest match between stored n-grams and the current. 
With $g^{(n)}$ as the last $n$
tokens of the current context, let
\begin{equation}
  n^\star=\max\{\,n\le N : C_n(g^{(n)},\cdot)\ \text{is nonempty}\,\}
  \label{eq:order}
\end{equation}
be the highest order at which the trailing context has been seen. The
predictive distribution of the next token is the normalized decayed counts at that single
order,
\begin{equation}
  p(t)\;=\;\frac{C_{n^\star}\!\big(g^{(n^\star)},t\big)}
                {\sum_{t'} C_{n^\star}\!\big(g^{(n^\star)},t'\big)},
  \label{eq:pred}
\end{equation}
and the drafted token is $\arg\max_t p(t)$. Orders are not interpolated:
the single highest order is used as draft token; lower orders are ignored. If
no order fires the store abstains.

\paragraph{Confidence.}
Each prediction also carries a scalar confidence $\kappa\in[0,1]$ that the
gate (Section~\ref{sec:gatedef}) thresholds. $\kappa$ is heuristically defined as the 
match-order score times one minus the normalized entropy of $p$,
\begin{equation}
  \kappa \;=\; \frac{n^\star}{N}\,
  \Big(1-\tfrac{H(p)}{\log|S|}\Big),
  \label{eq:conf}
\end{equation}
where $H(p)=-\sum_{t\in S} p(t)\log p(t)$ is the entropy of $p$ over its
support $S=\{t:p(t)>0\}$, the ratio $H(p)/\log|S|\in[0,1]$ normalizes it by
its maximum, and a singleton support ($|S|=1$) takes the entropy term to
be $0$. The match-order score $n^\star/N$ takes the values
$\{1,\tfrac34,\tfrac12,\tfrac14\}$ for orders $\{4,3,2,1\}$. Confidence is
thus highest for a long deterministic match (order $4$, single successor,
$\kappa=1$) and lowest for a short high-entropy one (order $1$, near-uniform
successors, $\kappa\to0$). The score defines $\kappa$, not $p$: it governs
whether the prediction is used, and not which token is drafted.

\paragraph{Statelessness and cost.}
The store is rebuilt from the context at the start of each draft call by one
left-to-right pass (Eq.~\ref{eq:counts}) in $O(TN)$ time with hashed lookups,
and holds no state between calls. It learns only from verifier-accepted
tokens: the $k$ tokens drafted within a call extend the query context
$g^{(n)}$ for later positions of that same draft but are never added to
$C_n$, so a rejected draft leaves no trace.
Its cost grows with the context: 280\,\textmu s per call at 100 generated
tokens and 600\,\textmu s at 500, against 10--14\,\textmu s for prompt
lookup (Section~\ref{sec:cost}).

\paragraph{How it differs from prompt lookup.}
Three differences separate the two. First, aggregation: lookup copies the
continuation of one occurrence, while session memory pools every
occurrence of the matched context into a distribution, discounted by
distance (Eq.~\ref{eq:counts}), and drafts its most probable successor.
Second, the drafting unit: lookup copies a contiguous span of $k$ tokens
from one place and needs $k$ tokens of continuation to exist, while
session memory re-predicts each position from the extended context, so a
draft can splice continuations from different places and can run past the
end of any single match. Third, reach: session memory matches contexts of
up to four tokens, lookup up to three. Both back off to shorter contexts
and both abstain when nothing matches; gating is an option for either
(Section~\ref{sec:gatedef}).

\subsection{The gate}
\label{sec:gatedef}

A threshold $\beta$ on the confidence $\kappa$ turns session memory into a
selective drafter. At each draft position, if $\kappa\ge\beta$ the store's
argmax is drafted; otherwise the draft stops there (\emph{gated session
memory}). Without a threshold ($\beta$=0), session memory drafts whenever
any order matches. We use $\beta$=0.5, the value of the first version of
this work.

The threshold has a simple reading. With $N$=4 and $\beta$=0.5,
$\kappa\ge\beta$ requires $n^\star\ge 2$ (Eq.~\ref{eq:conf}); at
$n^\star$=2 it additionally requires a single observed successor, at
$n^\star$=3 a normalized entropy of at most $1/3$, and at $n^\star$=4 at
most $1/2$. The gate is therefore an order floor of two, tightened by a
consistency requirement that relaxes as the matched context grows.
Section~\ref{sec:energy} tests which half of that rule does the work.

\section{Experimental Setup}
\label{sec:setup}

\paragraph{Verifier and hardware.}
Mistral-7B-v0.1~\citep{mistral2023}\footnote{\url{https://huggingface.co/mistralai/Mistral-7B-v0.1}}
in \texttt{float16} decodes greedily on the MPS accelerator of a Mac Studio
(Apple M1 Max, 32\,GB RAM), with the persistent KV cache and rejection
fast-path of Section~\ref{sec:system}. All drafters run on CPU.

\paragraph{Prompts.}
We evaluate on the test splits of WikiText-103~\citep{merity2017wikitext}
(general), PubMed Summarization~\citep{cohan2018pubmed} (medical), and
EurLex~\citep{chalkidis2022lexglue} (legal). From each split we take, in its
native order, the first 100 documents of at least 60 words and use each
document's first 50 whitespace-delimited words as a prompt, which averages
67, 72, and 100 Mistral tokens on general, medical, and legal
text.\footnote{\url{https://huggingface.co/datasets/Salesforce/wikitext},
\url{https://huggingface.co/datasets/ccdv/pubmed-summarization},
\url{https://huggingface.co/datasets/coastalcph/lex_glue} (\texttt{eurlex}).}
The verifier generates 100 tokens unless stated otherwise.

\paragraph{Experiments.}
The prompt-lookup comparison (Section~\ref{sec:pld}) runs plain decoding,
both lookup forms, and ungated session memory on all 300 prompts at
$k\in\{3,5,8\}$. The 500-token comparison (Section~\ref{sec:pld500}) uses
the first 30 general and legal prompts at $k$=5 and adds gated session
memory. The energy sweeps (Section~\ref{sec:energy}) repeat both with
system power logged, and add the gated lookup forms. The within-generation
analysis (Section~\ref{sec:curve}) uses the per-step traces of the
500-token runs.

\paragraph{Protocol.}
Each domain's sweep starts with plain decoding of its prompts in the same
session, and speedup is the per-prompt ratio of a drafter's throughput to
that baseline. Differences between drafters are paired over prompts, with
95\% percentile intervals from 10{,}000 bootstrap resamples of prompts.
Ungated session memory, re-run alongside the new drafters, reproduces
the first version of this work within 0.2 points of acceptance in all nine
configurations and within 0.03 of speedup at $k\le5$ (0.05 at $k$=8;
per-prompt correlation 0.97), although operating-system and library
updates since lowered absolute throughput by about 23\%; we therefore
compare only ratios measured within one session.

\paragraph{Metrics.}
We report acceptance (accepted over proposed draft tokens), tokens per
verifier step (generated tokens over decoding steps, which sets
throughput), abstention (the share of draft calls proposing nothing),
drafting time as a share of wall time, energy per 100 tokens
(Section~\ref{sec:energy}), and exact token agreement with greedy decoding
(Section~\ref{sec:agree}).

\section{Speed: Recency, Not Weighting}
\label{sec:results}
\label{sec:speed}

\begin{table*}[t]
\centering
\small
\begin{tabular}{llrrrcc}
\toprule
 & & \multicolumn{3}{c}{Speedup over plain decoding} & \multicolumn{2}{c}{Difference} \\
\cmidrule(lr){3-5} \cmidrule(lr){6-7}
Domain & $k$ & Lookup, earliest & Lookup, most recent & Session memory & recency & weighting \\
\midrule
General & 3 & 1.153 {\scriptsize[1.122, 1.183]} & 1.169 {\scriptsize[1.139, 1.200]} & 1.165 {\scriptsize[1.134, 1.197]} & +0.016$^*$ & -0.004$^*$ \\
General & 5 & 1.216 {\scriptsize[1.170, 1.266]} & 1.236 {\scriptsize[1.187, 1.286]} & 1.227 {\scriptsize[1.179, 1.280]} & +0.020$^*$ & -0.009$^*$ \\
General & 8 & 1.217 {\scriptsize[1.159, 1.282]} & 1.245 {\scriptsize[1.181, 1.311]} & 1.243 {\scriptsize[1.177, 1.313]} & +0.027$^*$ & -0.002 \\
Medical & 3 & 1.110 {\scriptsize[1.089, 1.132]} & 1.120 {\scriptsize[1.100, 1.143]} & 1.116 {\scriptsize[1.095, 1.139]} & +0.011$^*$ & -0.004$^*$ \\
Medical & 5 & 1.139 {\scriptsize[1.106, 1.174]} & 1.152 {\scriptsize[1.119, 1.188]} & 1.148 {\scriptsize[1.115, 1.184]} & +0.013$^*$ & -0.004$^*$ \\
Medical & 8 & 1.127 {\scriptsize[1.085, 1.175]} & 1.139 {\scriptsize[1.095, 1.190]} & 1.131 {\scriptsize[1.087, 1.179]} & +0.012 & -0.008$^*$ \\
Legal & 3 & 1.130 {\scriptsize[1.115, 1.146]} & 1.128 {\scriptsize[1.113, 1.143]} & 1.123 {\scriptsize[1.108, 1.138]} & -0.002 & -0.005$^*$ \\
Legal & 5 & 1.147 {\scriptsize[1.125, 1.171]} & 1.143 {\scriptsize[1.123, 1.164]} & 1.135 {\scriptsize[1.116, 1.156]} & -0.003 & -0.008$^*$ \\
Legal & 8 & 1.141 {\scriptsize[1.113, 1.174]} & 1.134 {\scriptsize[1.106, 1.165]} & 1.122 {\scriptsize[1.098, 1.148]} & -0.007 & -0.012$^*$ \\
\bottomrule
\end{tabular}
\caption{Prompt lookup decoding versus session memory, KV-cached Mistral-7B, 100-token
generations, 100 prompts per domain. Speedup is the per-prompt ratio to plain decoding in
the same session, mean with 95\% bootstrap interval over prompts. \emph{recency} is
most-recent minus earliest match (what recency alone buys); \emph{weighting} is session
memory minus most-recent lookup (what recency weighting and backoff add on top).
$^*$: interval excludes zero.}
\label{tab:pld}
\end{table*}
 
\subsection{Prompt lookup versus session memory}
\label{sec:pld}

Table~\ref{tab:pld} compares the three drafters on 100-token generations.
All lie within 0.03 of each other, between 1.10$\times$ and 1.25$\times$,
while per-prompt speedups spread widely (standard deviations near 0.25 on
general text), so we read differences from paired intervals.

Recency helps where text repeats. Most-recent lookup beats earliest-match
lookup by $+0.016$ to $+0.027$ on general text at every draft length and
by about $+0.01$ on medical text, and makes no difference on legal text,
where early and late occurrences of formulaic contexts tend to continue
identically. Weighting and backoff do not help. Session memory is below
most-recent lookup in all nine cells, by 0.002 to 0.012, and never reliably
above it. Its tokens per verifier step match lookup's to within 0.02 (for
example 1.743 against 1.740 on general text at $k$=5), so the small deficit
is its drafting cost, 0.28\% of wall time against 0.01\%.

\begin{table*}[t]
\centering
\small
\begin{tabular}{llrrrcr}
\toprule
Domain & Drafter & Acc.\ \% & Abstain \% & Tok./step & Speedup & Draft \% wall \\
\midrule
General & Lookup, earliest & 49.9 & 30 & 2.82 & 1.622 {\scriptsize[1.500, 1.740]} & 0.01 \\
General & Lookup, most recent & 56.3 & 32 & 2.98 & 1.711 {\scriptsize[1.585, 1.829]} & 0.01 \\
General & Session memory & 55.4 & 31 & 2.98 & 1.703 {\scriptsize[1.578, 1.821]} & 0.39 \\
General & Session memory, gated & 76.4 & 55 & 2.81 & 1.682 {\scriptsize[1.550, 1.806]} & 0.41 \\
\midrule
Legal & Lookup, earliest & 30.7 & 28 & 2.11 & 1.366 {\scriptsize[1.316, 1.419]} & 0.01 \\
Legal & Lookup, most recent & 35.9 & 30 & 2.27 & 1.450 {\scriptsize[1.384, 1.517]} & 0.01 \\
Legal & Session memory & 36.1 & 30 & 2.28 & 1.435 {\scriptsize[1.368, 1.503]} & 0.50 \\
Legal & Session memory, gated & 60.9 & 66 & 2.01 & 1.406 {\scriptsize[1.346, 1.467]} & 0.56 \\
\bottomrule
\end{tabular}
\caption{500-token generations, $k=5$, 30 prompts per domain. Acceptance is per
proposed token; \emph{abstain} is the share of draft calls that proposed nothing;
tokens per step counts verifier calls. The gate raises acceptance by abstaining, which
lowers tokens per step and speed.}
\label{tab:pld500}
\end{table*}
 
\subsection{Long generations}
\label{sec:pld500}

Over 500-token generations (Table~\ref{tab:pld500}) the recency effect
grows about fourfold: most-recent lookup is $+0.084$ [$+0.060$, $+0.111$]
faster than earliest-match lookup on legal text, winning on all 30
prompts, and $+0.089$ [$+0.055$, $+0.126$] on general text. As a
generation lengthens, the earliest occurrence of a context is increasingly
likely to predate the text's current phrasing, while the most recent one
is where the verifier is repeating itself now. Session memory still adds
nothing: ungated, it is $-0.015$ [$-0.029$, $+0.003$] below most-recent
lookup on legal text and $-0.009$ [$-0.016$, $-0.003$] on general text,
with the same tokens per step.

The gate raises acceptance from 36\% to 61\% on legal text and from 55\%
to 76\% on general text, but it does so by abstaining on 55--66\% of draft
calls against 30\% for the ungated store. Fewer drafts mean fewer tokens
per verifier step (2.01 against 2.28 on legal text), and the gated store is
0.02--0.03 slower than the ungated one. When drafters abstain at different
rates, acceptance per proposed token does not measure drafting value;
tokens per verifier step does.

\subsection{Drafting cost}
\label{sec:cost}

Session memory rebuilds its store at every draft call in $O(TN)$ time, so
its cost per call grows with the context and its cumulative cost with the
square of the output length. We measure 280\,\textmu s per call at 100
generated tokens (0.28\% of wall time) and 600\,\textmu s at 500 tokens
(0.45--0.48\%). Our unoptimized Python prompt lookup costs 10 and
14\,\textmu s (0.01\%). Both are small against a verifier step of about
75\,ms at these lengths, but session memory's share would reach several
percent at a few thousand tokens without an incremental, $O(N)$ per
token update, which we leave to future work.

\begin{table*}[t]
\centering
\small
\begin{tabular}{lrrrrrr}
\toprule
 & \multicolumn{3}{c}{Legal} & \multicolumn{3}{c}{General} \\
\cmidrule(lr){2-4} \cmidrule(lr){5-7}
 & \multicolumn{2}{c}{Speedup} & J/100 & \multicolumn{2}{c}{Speedup} & J/100 \\
Drafter & greedy & penalised & penalised & greedy & penalised & penalised \\
\midrule
Plain decoding & 1 & 1 & 106.4 & 1 & 1 & 101.7 \\
Lookup, earliest & 1.37 & 1.09 & 107.7 & 1.62 & 1.14 & 99.9 \\
Lookup, most recent & 1.45 & 1.12 & 105.9 & 1.71 & 1.16 & 98.7 \\
Lookup, 3-token floor & 1.40 & 1.09 & 102.0$^*$ & 1.67 & 1.16 & 95.3$^*$ \\
Session memory, gated & 1.41 & 1.08 & 105.9 & 1.68 & 1.14 & 96.5$^*$ \\
\bottomrule
\end{tabular}
\caption{500-token generations with a repetition penalty ($\theta$=1.2) on the verifier,
$k=5$, 30 prompts per domain, against greedy decoding. The penalty lowers the share of
positions in the last 100 tokens whose preceding 4-gram occurred earlier from 81\% to
28\% (legal) and from 84\% to 44\% (general). Energy in joules per 100 tokens;
$^*$: paired saving over plain decoding with a 95\% bootstrap interval excluding zero.}
\label{tab:rep}
\end{table*}
 
\subsection{Within-generation dynamics and repetition}
\label{sec:curve}
\label{sec:repetition}

\begin{table}[t]
\centering
\small
\setlength{\tabcolsep}{3pt}
\resizebox{\columnwidth}{!}{%
\begin{tabular}{llrrrrr}
\toprule
 & & \multicolumn{5}{c}{Generated tokens} \\
\cmidrule(lr){3-7}
 & Drafter & 1--100 & --200 & --300 & --400 & --500 \\
\midrule
Legal & Lookup, earliest & 0.41 & 0.79 & 1.34 & 1.62 & 2.22 \\
Legal & Lookup, most recent & 0.43 & 0.87 & 1.50 & 1.92 & 2.68 \\
Legal & Session memory, gated & 0.30 & 0.62 & 1.24 & 1.60 & 2.29 \\
\midrule
General & Lookup, earliest & 0.44 & 1.55 & 2.06 & 2.07 & 2.34 \\
General & Lookup, most recent & 0.44 & 1.77 & 2.52 & 2.39 & 2.79 \\
General & Session memory, gated & 0.34 & 1.47 & 2.14 & 2.04 & 2.59 \\
\bottomrule
\end{tabular}}
\caption{Accepted draft tokens per verifier step by position within 500-token greedy
generations ($k=5$, 30 prompts per domain). Drafting value rises as the output
accumulates, and the gap between most-recent and earliest-match lookup widens.}
\label{tab:curve}
\end{table}
 
Session-local drafters learn from the generation as it accumulates.
Table~\ref{tab:curve} tracks accepted draft tokens per verifier step over
500-token greedy generations: on legal text most-recent lookup rises from
0.43 in the first hundred tokens to 2.68 in the last, and gated session
memory's acceptance per proposed token climbs from 34\% to 78\% (47\% to
86\% on general text). The gap between the two lookup forms widens with
position, from 0.02 to 0.46 tokens per step on legal text, which is where
the fourfold growth of the recency effect between 100 and 500 tokens comes
from. Much of the climb is the verifier repeating itself, a known property
of maximisation-based decoding~\citep{holtzman2020curious}: positions
whose preceding 4-gram occurred earlier in the output rise from 7\% of
the first hundred tokens to 81\% of the last on legal text, which is also
why recency-aware copying captures the effect as well as the full store
does.

\paragraph{Without the repetition.} To test whether the gains survive
text that does not repeat itself, we rerun the 500-token comparison with
the repetition penalty of \citet{keskar2019ctrl} ($\theta$=1.2) on the
verifier. Verification stays exact: each draft position is checked
against the penalised argmax given the context up to it, and penalised
plain decoding reproduces the library's penalised generation. The penalty works as intended (Table~\ref{tab:rep}): in the last
100 tokens, repeated 4-gram contexts fall from 81\% to 28\% on legal and
from 84\% to 44\% on general text, distinct bigrams rise from 0.39 to 0.70
and from 0.25 to 0.72, and perplexity under the unpenalised verifier stays
low (1.58 and 2.35, against 1.19 and 1.30 for the degenerate greedy
output); 11--12 of 30 outputs end before 500 tokens. Without the loops,
every drafter falls to 1.08--1.16$\times$: most of the long-generation
gain was the verifier repeating itself. Recency still helps, with
most-recent lookup ahead of earliest-match lookup by $+0.029$ in both
domains (legal [$+0.018$, $+0.044$], general [$-0.000$, $+0.056$]), and
session memory still adds nothing over it.

\section{Energy: Selectivity, Not Speed}
\label{sec:energy}

\begin{table*}[t]
\centering
\small
\begin{tabular}{lrrrrrr}
\toprule
 & & \multicolumn{3}{c}{100 tokens, $k=5$} & \multicolumn{2}{c}{500 tokens, $k=5$} \\
\cmidrule(lr){3-5} \cmidrule(lr){6-7}
Drafter & Abstain \% & General & Medical & Legal & General & Legal \\
\midrule
Plain decoding & -- & 92.7 & 93.7 & 98.2 & 105.3 & 108.2 \\
Lookup, most recent & 41 & 89.5$^*$ & 94.6$^*$ & 99.2$^*$ & 78.2 & 90.0$^*$ \\
Lookup, 2-token floor & 69 & 87.5$^*$ & 92.0$^*$ & 96.8$^*$ & 77.7 & 88.4 \\
Lookup, 3-token floor & 79 & 86.7 & 91.2 & 96.1$^*$ & 78.4 & 88.6 \\
Lookup, agreement & 64 & 88.6$^*$ & 93.4$^*$ & 96.7$^*$ & 78.7$^*$ & 89.7$^*$ \\
Session memory, gated & 72 & 86.8 & 91.0 & 95.3 & 77.5 & 87.9 \\
\bottomrule
\end{tabular}
\caption{Energy in joules per 100 generated tokens (CPU+GPU+ANE power integrated over
each generation; mean over prompts). \emph{Abstain} is the mean share of draft calls
that proposed nothing. $^*$: paired difference from gated session memory has a 95\%
bootstrap interval excluding zero. A 3-token order floor on lookup comes within 1\,J of gated
session memory in every column.}
\label{tab:energy}
\end{table*}
 
Energy per token is not speed inverted. \citet{vandenbosch2026green} showed
that under a cached verifier mean system power rises with draft length,
because dense $k$-token verification passes work the accelerator harder than
one-token steps, so a faster configuration can cost more energy. We measure
whole-system power (CPU, GPU and neural engine) at 1\,Hz with
\texttt{powermetrics}, integrate it over each generation, and report joules
per 100 generated tokens, paired per prompt against plain decoding in the
same session. Table~\ref{tab:energy} covers draft length 5, where the gate's
effect on speed is largest, at 100 tokens on all three domains and at 500
tokens on two.

\paragraph{Unselective drafting saves time, not energy.} On medical and
legal text at 100 tokens, most-recent lookup is 1.14$\times$ faster than
plain decoding and uses 94.6 and 99.2\,J per 100 tokens against plain's
93.7 and 98.2: no saving at all. Mean power rises from 12.5--12.7\,W for
plain to 14.2--14.6\,W, and the rise eats the time saved. Ungated session
memory behaves the same way. Only on general text, where acceptance is
higher, does an unselective drafter save energy (89.5 versus 92.7\,J).

\paragraph{The gate converts speed into energy.} Gated session memory has
the lowest energy in every column of Table~\ref{tab:energy}. Against
most-recent lookup it saves 2.7, 3.6 and 3.9\,J per 100 tokens on general,
medical and legal text, and 2.1\,J on 500-token legal generations, with
bootstrap intervals well clear of zero; on 500-token general text the
difference is within noise. It pays for this with $0.02$--$0.03$ of speedup
(Section~\ref{sec:speed}). The mechanism is visible in the abstention
column: the gate declines to draft on 78--82\% of calls at 100 tokens
and 55--66\% at 500, and the tokens it declines to draft are the ones a verifier would most
likely have rejected. Verifying a rejected draft costs a dense forward
pass and returns one token, which is the expensive case; abstaining costs
a cheap single-token pass and returns one token.

\paragraph{An order floor on lookup does the same job.} If abstention is
the mechanism, prompt lookup should gain as much from a rule that declines
short matches. It does. Drafting only on matches of two or more tokens
recovers about three quarters of the gate's saving; drafting only on
matches of three tokens matches gated session memory within noise in all
five columns (the paired difference is at most 0.9\,J and its interval
includes zero in four of five). Consistent with the reading of the gate
in Section~\ref{sec:gatedef}, an order floor of two tightened by a
consistency requirement, the lookup rule that matches it best is the
stricter floor of three. A gate on the \emph{agreement} of earlier
continuations, the closest lookup analogue of session memory's entropy
term, abstains as often but saves less, so what matters is the length of
the matched context, not the consensus of what followed it. The relation
is monotone: across the six drafters, more abstention means less energy,
at a speed cost that never exceeds $0.03$. At draft length 3 the effect
shrinks to under 1\,J per 100 tokens, as expected when a rejected step
wastes less.

\paragraph{Without repetition.} Under the repetition penalty
(Table~\ref{tab:rep}), unselective lookup saves no energy at all (105.9
against 106.4\,J per 100 tokens on legal text, 98.7 against 101.7 on
general text, intervals including zero). The three-token floor is the one
drafter that saves reliably in both domains, 4.4\,J [2.6, 6.2] on legal
and 6.5\,J [2.2, 11.2] on general text, and on legal text it beats gated
session memory by 3.9\,J [2.0, 6.2], while the gate's own saving there is
not measurable.

\paragraph{Recommendation.} Under a cached verifier, draft with
session-local lookup from the most recent match and refuse contexts
shorter than three tokens: no parameters, no training, 10--14\,\textmu s
per call, the energy saving of the gated session memory without its store,
and the only saving that holds whether or not the output repeats itself.

\section{Output Agreement}
\label{sec:agree}

Acceptance compares token IDs with the verifier's greedy choice, so
speculative decoding should reproduce greedy output exactly. It need not
quite, because a $k$-token verification pass and $k$ single-token passes
compute marginally different \texttt{float16} logits. We measure the
consequence directly on the 100-token generations of
Section~\ref{sec:pld} (breakdown in Table~\ref{tab:agree},
Appendix~\ref{app:agree}). Two non-speculative
references, the library's greedy generation function and our decoder with
drafting disabled, agree on all 300 prompts. Against them, 2{,}539 of
2{,}700 speculative generations (94.0\%) are token-identical. Agreement
does not depend on the drafter (93.4--94.4\%) or the draft length
(93.8--94.2\%), and the 161 divergent generations reduce to 47 distinct
divergence points in 45 prompts: the same prompt diverges at the same
position, to the same token, whichever drafter runs. The first divergence
falls at a median of token 38; two thirds of divergent sequences share an
8-token run with the reference again later, and the rest stay split.

Every divergence is a numerical tie. Teacher-forcing the greedy reference
through the verifier, the gap between the top two logits at the 47
divergence points has a median of 0 (25 are exact \texttt{float16} ties)
and a maximum of 0.023, against a median of 3.0 over all 30{,}000
generated positions; only 1.1\% of positions have a gap that small. In 45
of the 47 cases the speculative token is the reference's runner-up. A
\texttt{float32} re-run, which would break these ties, does not fit in the
machine's 32\,GB.

\section{Related Work}
\label{sec:related}

Speculative decoding was introduced with a smaller neural model as the
draft~\citep{leviathan2023speculative,chen2023speculative}. Trained
drafting heads attached to the verifier~\citep{cai2024medusa,li2024eagle},
with dynamic draft trees in EAGLE-2~\citep{li2024eagle2}, give the
strongest speedups on accelerators but must be trained per verifier and
add accelerator work, which places them outside the training-free,
CPU-only setting studied here. Training-free drafters retrieve
continuations from a datastore~\citep{he2023rest}, a reference
document~\citep{yang2023inference}, the prompt~\citep{saxena2023prompt},
or suffix structures over the prompt and earlier
outputs~\citep{oliaro2025suffix,hu2024sam}. Our
results support that design direction: what separates good drafts from
wasted verification is the recency and length of the matched context, not
a weighted distribution over its continuations. Memory-based language
models~\citep{daelemans2005mbl,vandenbosch2025olifant} provide a trained,
CPU-resident draft at similar cost; \citet{vandenbosch2026green} evaluate
them as static drafters, with energy under cached and uncached
verification.

That a text's recent history predicts its continuation is old knowledge:
words recur in bursts~\citep{churchgale1995poisson,church2000noriegas},
cache language models exploit this~\citep{kuhndemori1990cache,rosenfeld1996maxent},
and neural models retrieve recent context through caches, pointers, and
nearest neighbours~\citep{grave2017continuouscache,merity2017wikitext,khandelwal2020generalization}.
These target perplexity; session-local drafting uses the same device to cut
cost.

Online speculative decoding~\citep{liu2024onlinespec} adapts a neural
draft across queries by distillation; session-local drafting adapts within
one generation.

\section{Conclusion}
\label{sec:conclusion}

Session-local drafting suits a cached verifier on constrained hardware:
it needs no training, no accelerator, and almost no time, and having no
training data, it cannot meet the wrong domain. Of its ingredients, only
two matter. For speed, copy from the most recent occurrence of the
current context; recency weighting and backoff add nothing, at 30 to 40
times the drafting cost. For energy, refuse to draft on contexts shorter
than three tokens, which reproduces the gated session memory's saving and,
once the verifier's repetition is suppressed and speedups shrink to
1.1--1.2$\times$, is the only rule that still saves energy. Output
matches greedy decoding in 94\% of generations and differs elsewhere only
at numerical ties. The resulting drafter is a scan over the text so far
and a length check.

\section*{Limitations}

\paragraph{Scope.} All measurements use one verifier, Mistral-7B, on one
consumer platform. Verifiers with other attention designs, and batched
serving, shift the balance between drafting and verification cost that
drives both our speed and our energy results. Decoding is greedy, with
and without a repetition penalty; sampling changes the acceptance rule and
is untested. The penalty shows that long-generation speedups depend
heavily on the verifier's self-repetition (Section~\ref{sec:curve}), and
we test a single penalty strength. We measure 100- and 500-token outputs from
prompts cut at a fixed 50-word length; longer contexts, conversational and
code workloads, and non-English text are untested.

\paragraph{Implementation.} Session memory rebuilds its store at every
call, with cumulative cost quadratic in output length
(Section~\ref{sec:cost}); an incremental store would remove this but we
did not implement it. The prompt-lookup forms are straightforward Python,
not optimized; since they are already 30 to 40 times cheaper per call,
optimization would only widen the gap.

\paragraph{Measurement.} Energy figures are whole-system and
hardware-specific, and support relative comparison on identical hardware,
not absolute cross-platform claims. Background load and backup activity were logged every 30\,s during all
timed runs, and blocks that overlapped load bursts were re-run.
Divergences from greedy output were characterized by logit margins, not
by a \texttt{float32} re-run.

Code and scripts to reproduce every table are
available at \url{https://github.com/antalvdb/olifant-speculative}.

\section*{Ethical Considerations}

This work targets the efficiency of language-model inference, and its main
impact is a reduction in the energy cost of on-device text generation: the
selective drafters we recommend use 2--7\% less energy per token than
plain decoding at 100 tokens, 18--26\% less at 500 tokens of greedy
output, and 4--6\% less at 500 tokens under a repetition penalty. Speculative
output is token-identical to greedy decoding in 94\% of generations and
differs elsewhere only where the verifier's top candidates are numerically
tied, so the methods alter the cost of generation, not its content, and
introduce no capability or misuse surface beyond that of the underlying
verifier. Session-local drafters store only tokens the verifier has itself
already produced, adding no external data and no new privacy exposure; by
keeping drafting on CPU with no accelerator footprint, they support fully
on-device generation in privacy-sensitive settings such as the medical and
legal domains we study. Two caveats apply: cheaper inference can raise
aggregate usage, a rebound effect that may partially offset per-token
savings, and our energy figures are whole-system and hardware-specific.

\bibliography{olifant_arr}

\appendix

\section{Token Agreement by Domain, Drafter, and Draft Length}
\label{app:agree}

\begin{table}[t]
\centering
\small
\begin{tabular}{llrr}
\toprule
 & & Identical & \% \\
\midrule
Domain & General & 840/900 & 93.3 \\
Domain & Medical & 838/900 & 93.1 \\
Domain & Legal & 861/900 & 95.7 \\
\midrule
Drafter & Lookup, earliest & 841/900 & 93.4 \\
Drafter & Lookup, most recent & 850/900 & 94.4 \\
Drafter & Session memory & 848/900 & 94.2 \\
\midrule
$k$ & 3 & 847/900 & 94.1 \\
$k$ & 5 & 844/900 & 93.8 \\
$k$ & 8 & 848/900 & 94.2 \\
\bottomrule
\end{tabular}
\caption{Exact token agreement between speculative output and greedy decoding
(2{,}700 100-token generations). The 47 distinct divergence
points have a top-two logit margin of at most 0.023
(median 0.0) against a median of
3.03 over all generated positions; in
45 of 47 the speculative
token is the reference's runner-up.}
\label{tab:agree}
\end{table}
 
\end{document}